\documentclass[runningheads]{llncs}

\usepackage{eccv}

\usepackage{eccvabbrv}

\usepackage{graphicx}
\usepackage{booktabs}

\usepackage[accsupp]{axessibility}  

\usepackage{hyperref}

\usepackage{orcidlink}

\usepackage{multirow}
\usepackage{xcolor}
\usepackage{colortbl, booktabs}
\usepackage{comment}
\usepackage[most]{tcolorbox}
\usepackage{siunitx}
\usepackage{amssymb, pifont}
\usepackage{wasysym}
\usepackage{footnote}
\usepackage{threeparttable}

\definecolor{eccvblue}{rgb}{0.12,0.49,0.85}
\definecolor{RoseRed}{rgb}{1.00, 0.38, 0.39}
\definecolor{GrassGreen}{HTML}{74cc00}
\definecolor{EggYellow}{HTML}{ffd21f}
\definecolor{SkyBlue}{HTML}{38b6ff}
\definecolor{FlowerPurple}{HTML}{bea1f7}

\begin{document}

\title{DH-VLM: Dual-Horizon Cooperative Latent Reasoning for Autonomous Driving} 

\titlerunning{Dual-Horizon Cooperative Latent Reasoning for Autonomous Driving}

\author{Ziyi Song\inst{1}\orcidlink{0009-0004-1332-8344} \and
Chen Xia\inst{1}\orcidlink{0009-0008-1251-2712} \and
Hang Yu\inst{1}\orcidlink{0009-0008-4234-5118} \and
Sheng Zhou\inst{1, 2}\thanks{Corresponding author.}\and
Zhisheng Niu\inst{1}
}

\authorrunning{Z.~Song et al.}

\institute{Department of Electronic Engineering, Tsinghua University \and
State Key Laboratory of Intelligent Green Vehicle and Mobility, Tsinghua University \\
\email{\{songzy24,xiac23,yuhang22\}@mails.tsinghua.edu.cn}\\
\email{sheng.zhou@tsinghua.edu.cn, niuzhs@tsinghua.edu.cn}}

\maketitle

\begin{abstract}
  Large-scale language models for autonomous driving enable enhanced global understanding and long-horizon planning. However, when deployed in isolated vehicles, limited sensing range and occlusions restrict reliable decision-making, and the substantial computational and latency overhead makes on-board deployment impractical. Cooperative driving provides a potential solution by leveraging external agents for information exchange, but existing methods remain limited in semantic reasoning capability under practical constraints. To address these challenges, we propose DH-VLM, a dual-horizon cooperative latent reasoning framework that enables asymmetric semantic cooperation between the infrastructure and ego vehicle. The infrastructure aggregates multi-layer hidden states to form a global-reasoning horizon latent guidance, which is integrated into the ego model through an Infrastructure-Driven Latent Evolution mechanism for conditional latent refinement. This enables the ego vehicle to leverage long-range contextual understanding while preserving autonomous decision-making within its local planning horizon. Furthermore, we construct a cooperation-oriented question–answer (QA) dataset covering fundamental scene understanding and ego-personalized comprehension to support counterfactual and safety-aware reasoning. Extensive experiments demonstrate that DH-VLM achieves state-of-the-art planning performance, outperforming the previous state of the art by 14.6\% in L2 error and 26.9\% in collision rate. Compared with query-based end-to-end cooperative driving methods, our approach reduces the communication cost by 57.3\% and GPU memory usage by 25.5\%, while maintaining strong robustness against infrastructure guidance errors, providing a practical and robust paradigm for cooperative autonomous driving.

\end{abstract}

\section{Introduction}
\label{sec:intro}

\begin{figure} [tb]
    \centering \includegraphics[width=1.0\linewidth]{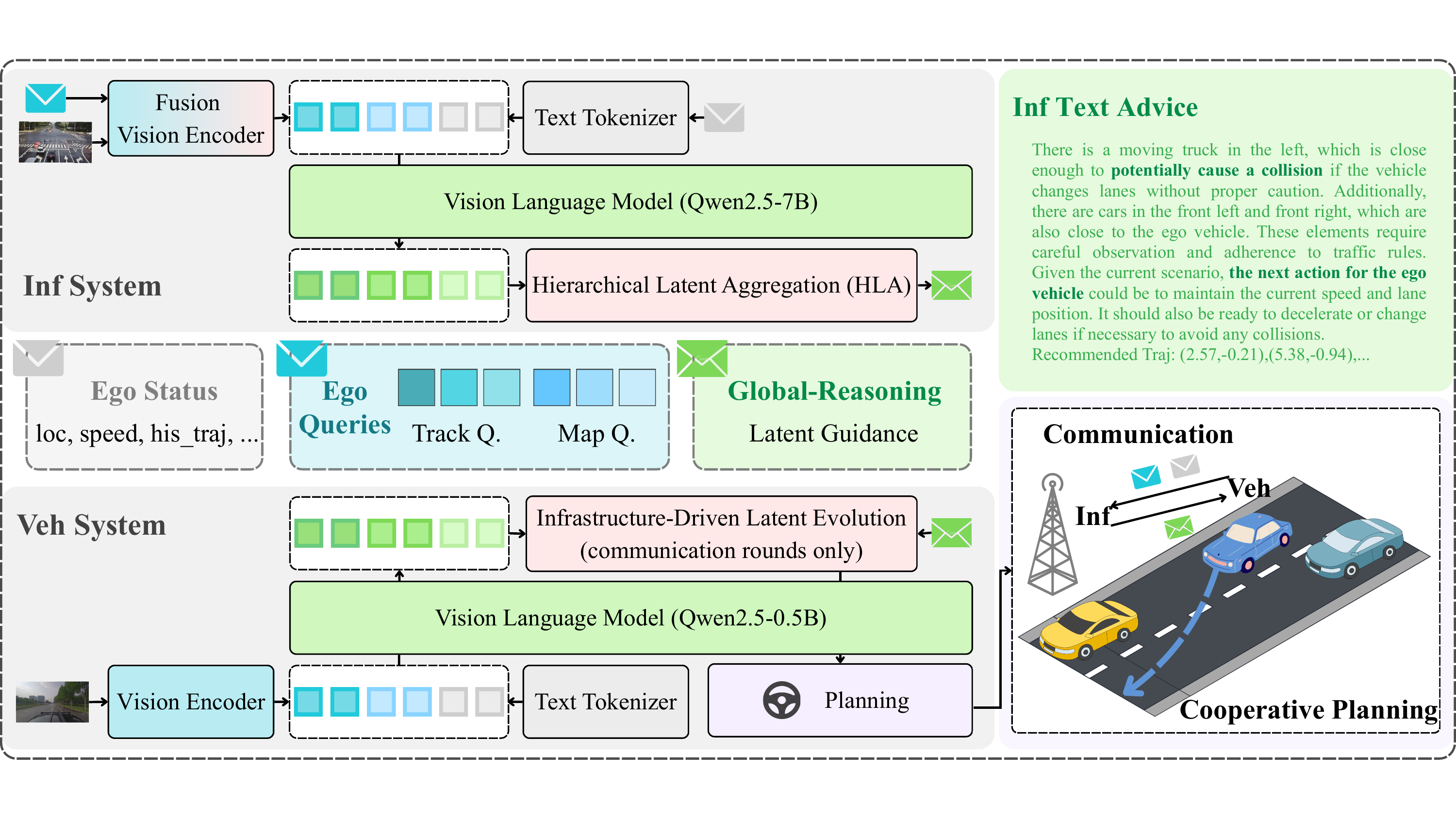} \caption{The overview of DH-VLM. The infrastructure performs global-reasoning horizon understanding by aggregating multi-agent perception queries through a fusion vision encoder, followed by an HLA module to compress multi-layer hidden states into a compact latent cognitive guidance. This information is transmitted via V2X communication to the ego vehicle, where an IDLE module conditionally refines ego hidden states, enabling cooperative latent-level reasoning while preserving local autonomy.} 
    \label{fig:framework} 
\end{figure}

Traditional autonomous driving has long been relying on modular pipelines, where perception, prediction, and planning are optimized independently. While modularity improves interpretability and engineering robustness, it suffers from error accumulation and information bottlenecks across stages. End-to-end autonomous driving addresses these limitations by enabling differentiable learning from raw sensory observations to control commands, demonstrating the benefits of joint optimization across different stages~\cite{uniad, vad, sparsedrive, diffusiondrive}. Nevertheless, these methods remain limited in long-tail scenarios, open-world generalization, and semantic reasoning, as feature maps lack the high-level semantic and causal understanding required for safe decision-making in complex environments.

Vision-Language Models (VLMs) offer strong zero-shot generalization and semantic reasoning capabilities by leveraging massive pretraining corpora, with recent works~\cite{opendrivevla, openemma, omnidrive, drivelm, drivevlm, gpt-driver, futuresightdrive, autovla} demonstrating the potential of VLMs for autonomous driving. Beyond explicit action prediction, latent-space reasoning models high-level latent dynamics as an intermediate semantic abstraction between perception and control~\cite{coconut, nextlat}. Recent works~\cite{latentvla, colavla} leverage latent-based methods to mitigate compounding errors in autonomous driving, highlighting latent dynamics as a scalable and interpretable foundation for robust autonomous driving systems. However, these approaches typically operate in isolation with limited perception range, increasing safety risks under occlusion or sensor blind spots. In contrast, Vehicle-to-Everything (V2X)-enabled cooperative driving extends perceptual coverage and enhances safety through shared information.

Cooperative autonomous driving has evolved from cooperative perception~\cite{where2comm, when2com, who2com, how2comm, coalign, cobevt, cooper, fcooper} to modular end-to-end frameworks~\cite{coopernaut, univ2x, unimm-v2x} that investigate how cross-agent representation fusion, such as BEV features and queries, influences planning performance. More recent works~\cite{v2x-vlm, langcoop} further incorporate VLMs into cooperative driving. Despite their potential, \cite{v2x-vlm} incurs substantial bandwidth overhead due to raw sensory transmission, while \cite{langcoop} imposes considerable computational demands that may exceed the capabilities of resource-constrained vehicles. These limitations motivate a fundamental question: \emph{how can cooperative autonomous driving systems effectively balance semantic reasoning capacity, communication efficiency under on-board constraints?}

To address these challenges, inspired by dual-system paradigms~\cite{drivevlm, latentvla, vlp} in single-agent intelligence, we propose DH-VLM, a \textbf{Dual-Horizon cooperative latent reasoning VLM architecture} that explicitly decouples global-reasoning horizon understanding from ego local-planning horizon, where a computationally powerful infrastructure provides global semantic guidance to assist a resource-constrained ego vehicle in safety-aware decision-making. Specifically, we introduce a latent-level fusion mechanism, where the infrastructure generates aggregated, semantically structured latent guidance to assist ego decision-making. By performing cooperation at the latent level rather than exchanging raw sensory data or text outputs, DH-VLM enhances semantic expressiveness while reducing communication overhead and preserving decision-making autonomy under on-board computational and temporal constraints.

For latent guidance extraction and aggregation on the infrastructure side, we introduce a \emph{Hierarchical Latent Aggregation (HLA)} module to distill global semantic representations from intermediate VLM hidden states. Motivated by prior findings~\cite{li2022emergent, clip, blip} that Transformer latent spaces encode structured visual–linguistic concepts, we aggregate multi-layer hidden states via attention and interpret the aggregated result as a continuous semantic guidance that captures global-reasoning horizon dependencies and global context. To incorporate this guidance, the ego vehicle adopts a two-stage forward process. It first performs a standard forward pass on its observations to obtain an initial latent state. When the infrastructure guidance is available, the initial latent state is refined via \emph{Infrastructure-Driven Latent Evolution (IDLE)}, followed by a second forward pass based on the evolved representation. If the guidance is unavailable due to asynchronous updates or communication packet loss, the ego model relies on the initial pass alone. This design preserves decision autonomy while enabling cooperative semantic conditioning and naturally accommodates the asynchronous update rates inherent in the dual-system architecture.

Furthermore, we construct a cooperation-oriented QA dataset built upon existing cooperative driving benchmark~\cite{dair}. In contrast to generic scene-level QA formulations, our dataset is explicitly tailored for ego-centric reasoning, facilitating guidance that is conditioned on the ego vehicle’s state and driving intent. It covers both fundamental semantic understanding (e.g., scene description, traffic regulations, and object presence) and ego-personalized comprehension (e.g., in-view ego masking, counterfactual reasoning, and safety-aware planning). By integrating structured QA annotations into multi-agent observations, the dataset systematically enhances semantic modeling capacity and promotes safety-aware scene reasoning for cooperative driving.

Our main contributions are summarized as follows:
\begin{itemize}
    \item We propose DH-VLM, a dual-horizon cooperative latent reasoning framework, where a high-capacity infrastructure assists a resource-limited ego vehicle, enhancing decision quality under on-board efficiency constraints.

    \item To the best of our knowledge, we are the first to explore latent-level fusion in VLM-based cooperative autonomous driving, enabling cooperative semantic reasoning without compromising ego autonomy.

    \item We construct a cooperation-oriented QA dataset to facilitate cooperative reasoning. Extensive experiments show that DH-VLM achieves SOTA planning performance with efficient communication and low computational cost.

\end{itemize}

\section{Related Works}
\subsection{End-to-End Autonomous Driving}
Early end-to-end methods~\cite{roach, tcp} map raw sensor data directly to final control via imitation learning. To improve interpretability, \cite{uniad} jointly models perception, prediction, and planning in a transformer-based architecture; \cite{vad} introduces vectorized representations to improve planning consistency; \cite{sparsedrive} employs sparse query mechanisms to reduce computational cost. However, these methods still struggle with long-tail scenarios and lack semantic reasoning. VLM-based approaches introduce stronger semantic understanding and reasoning processes into planning via large-scale pretraining~\cite{openemma, opendrivevla, omnidrive, autovla}. Dual-system designs further decouple high-level reasoning from on-board driving~\cite{drivevlm, vlm-ad, vlp, latentvla}. However, these methods remain limited by sensing ranges and occlusion, necessitating the cooperative driving frameworks that integrate multi-agent information exchange to achieve complementary reasoning.

\subsection{Cooperative Autonomous Driving}
Cooperative autonomous driving leverages V2X communication to overcome the perceptual limits of individual vehicles. Early works focus on cooperative perception~\cite{fcooper, when2com, where2comm, how2comm, who2com, coalign, cobevt, v2x-vit, v2vnet}, designing fusion strategies to balance detection accuracy and communication cost. However, perception-oriented objectives are often misaligned with planning~\cite{zeng2019end}. End-to-end cooperative frameworks address this by directly mapping multi-agent observations to ego control via BEV or query fusion~\cite{coopernaut, univ2x, unimm-v2x}. Recent VLM-based methods~\cite{v2x-vlm, langcoop} introduce reasoning into cooperation. However, result-level fusion is sensitive to upstream errors and introduces heavy on-board computational burden~\cite{langcoop}, while early-level fusion incurs substantial communication overhead~\cite{v2x-vlm}. We propose a dual-horizon cooperative framework that assigns global-horizon reasoning to the infrastructure while preserving decision autonomy at the vehicle, enabling semantic guidance through lightweight latent integration.

\subsection{Latent Representation and Cooperation}
Latent representations have emerged as a compact yet semantically dense medium for reasoning and communication in foundation models. The latent transition paradigm, introduced in \cite{coconut}, enables continuous reasoning directly within the latent space, bypassing the need for discrete token decoding. In autonomous driving, a unified latent space for vision–action alignment has been explored to avoid auto-regressive language generation \cite{colavla}. Building upon this direction, latent representations are further compressed via knowledge distillation to facilitate on-board deployment \cite{latentvla}. Beyond single-agent systems, recent robotics study on latent-level cooperation~\cite{latent-coop} explores multi-agent coordination through the final layer of hidden states exchange rather than raw data or text to facilitate efficient cooperation. Motivated by these advances, our dual-system framework transmits compact latent-level information from infrastructure to the ego vehicle, avoiding the error propagation of text-based reasoning and the bandwidth overhead of raw data sharing, while supporting on-board execution.

\section{Method}

We introduce a dual-horizon cooperative latent reasoning system tailored for vehicle-infrastructure autonomous driving. To fully leverage the capabilities of vision–language models and explore efficient information exchange in this emerging paradigm, we study latent-level fusion between the infrastructure and ego systems (\cref{sec:inf} and \cref{sec:veh}). To support our framework, we further construct a cooperation-oriented QA dataset (\cref{sec:data}). Our approach integrates global semantic guidance while preserving ego autonomy, enabling robust and efficient cooperative driving.

\subsection{Overview: Dual-Horizon Cooperative Driving}
\label{sec:framework}

In DH-VLM, the vehicle performs ego-centric planning and control for on-board decision-making, while the infrastructure provides high-level, global-reasoning horizon semantic guidance under relaxed latency constraints. Instead of directly affecting control outputs, the infrastructure information is encoded as latent guidance and fused into the ego model at the latent level. This forms an asymmetric and loosely coupled cooperation paradigm in which the ego vehicle retains execution authority and the infrastructure provides semantic guidance.

\textbf{Infrastructure System for Global-Reasoning Horizon.}
The infrastructure focuses on global reasoning over complex traffic evolution. It aggregates its perception queries with transmitted ego queries via the Fusion Vision Encoder module, producing fused perception tokens $Q_A$ and $Q_L$. A unified 3D coordinate space is adopted to ensure spatial alignment with the ego vehicle. The fused tokens are processed by a large language model to generate:
(i) semantic latent guidance $z_t^{\text{inf}}$ encoding global traffic context and driving intent through an HLA module; and
(ii) text-based reasoning and advice $T_t^{\text{inf}}$ for interpretability.

\textbf{Vehicle System for Local-Planning Horizon.}
The ego vehicle conducts continuous self-contained decision making architecture comprising a vision encoder, a lightweight language model, and an IDLE module. The vision encoder follows the same architecture as the infrastructure encoder but omits fusion components, decoupling dual-system interactions. The lightweight language model processes vision and text tokens for understanding and planning. When infrastructure latent guidance is available, the IDLE module performs conditional latent refinement; otherwise, the vehicle system follows its native pipeline, ensuring uninterrupted autonomous operation.

\begin{figure}[tb]
    \centering
    \includegraphics[width=1.0\linewidth]{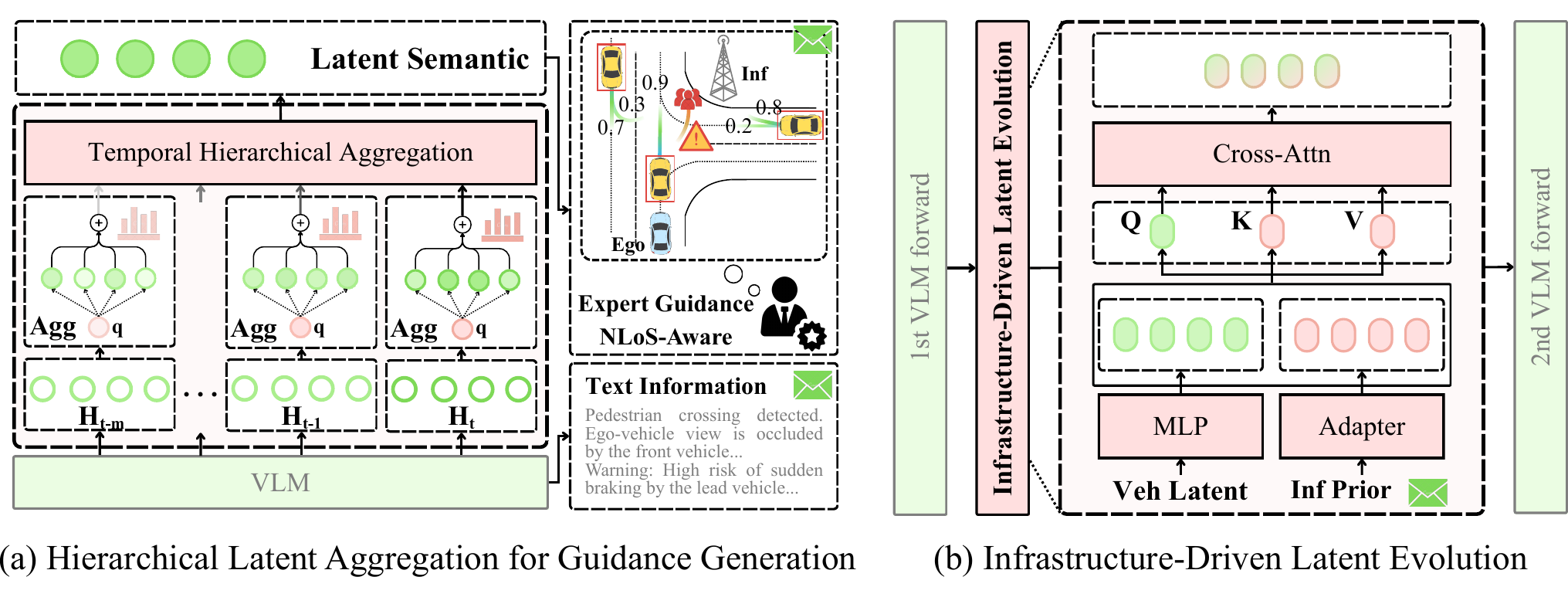}
    \caption{Latent-level interaction between the infrastructure and vehicle systems. (a) illustrates the hierarchical latent guidance generation in the infrastructure system, while (b) depicts how this guidance is utilized in the vehicle system via the proposed IDLE module with an attention-based mechanism.}
    \label{fig:fusion}
\end{figure}

\subsection{Hierarchical Latent Aggregation for Infrastructure Guidance}
\label{sec:inf}

Large language models encode hierarchical semantic abstractions of complex driving scenes through layered hidden representations.
Prior studies indicate that intermediate Transformer layers capture rich contextual, structural, and causal semantics, whereas the final layer is primarily optimized for token prediction and often compresses representations into compact decision-oriented embeddings.
To provide the ego vehicle with high-level semantic and causal guidance rather than deterministic decisions, we extract latent representations from intermediate reasoning layers of the infrastructure VLM, as shown in \cref{fig:fusion}(a).

Formally, let the VLM consist of $L$ Transformer layers with hidden states. We select a subset of intermediate layers $\mathcal{S} \subset \{1,\dots,L-1\}$ and treat their hidden states as a latent reasoning manifold:
\begin{equation}
\mathbf{Z}^{\text{inf}} = \{\mathbf{h}_l \mid l \in \mathcal{S}\}.
\end{equation}

To obtain a compact and robust infrastructure guidance, 
we propose an HLA module 
that consolidates hidden states across hierarchical dimensions. For each selected layer $l \in \mathcal{S}$, 
we perform attention-based semantic aggregation to summarize the latent states into a compact representation:
\begin{equation}
\text{Agg}(\mathbf{h}_l^{(t)}) 
= \sum_{i=1}^{N} \beta_{l,i}^{(t)} \mathbf{h}_{l,i}^{(t)}, \quad \beta_{l,i}^{(t)} =
\frac{\exp(\mathbf{q}^\top \mathbf{h}_{l,i}^{(t)})}
{\sum_{j=1}^{N} \exp(\mathbf{q}^\top \mathbf{h}_{l,j}^{(t)})},
\end{equation}
where $\mathbf{q}$ is a learnable parameter that selects salient tokens 
for global reasoning. We then aggregate across layers via learnable layer-wise weights:
\begin{equation}
\tilde{\mathbf{z}}_t^{\text{inf}}
= \sum_{l \in \mathcal{S}} \alpha_l \, \text{Agg}(\mathbf{h}_l^{(t)}),
\quad
\alpha_l =
\frac{\exp(w_l)}{\sum_{k \in \mathcal{S}} \exp(w_k)},
\end{equation}
where $w_l$ is a learnable parameter indicating 
the relative importance of layer $l$, allowing the model to adaptively weight semantic representations from different reasoning depths. To reduce temporal fluctuations and enhance stability, we further aggregate the hierarchical representations within a temporal window $\mathcal{T}$:
\begin{equation}
\mathbf{z}^{\text{inf}}_t = \mathcal{G}\left(\{\tilde{\mathbf{z}}_{t-\tau}^{\text{inf}}\}_{\tau \in \mathcal{T}}\right).
\end{equation}
This produces a temporally consistent latent representation that summarizes the infrastructure’s global-reasoning horizon understanding of the scene. The resulting $\mathbf{z}^{\text{inf}}_t$ is transmitted to the ego vehicle as a compact high-level cognitive latent guidance for cooperative reasoning and planning.

\subsection{Infrastructure-Driven Latent Evolution on the Vehicle}
\label{sec:veh}

We introduce the IDLE module that injects infrastructure latent guidance into the ego reasoning, enabling global scene understanding of occluded risks and distant traffic intent, while maintaining local autonomy and robustness.

As illustrated in \cref{fig:fusion}(b), at each time step $t$, the ego vehicle first extracts its semantic tokens $\tilde{Q}_t^{\text{veh}}$ by fusing on-board visual tokens $Q_t^{\text{veh}}$ and textual instructions $T_t$ through the first forward pass of the ego VLM:
\begin{equation}
\tilde{Q}_t^{\text{veh}} = \mathcal{F}_{\text{VLM}}^{\text{veh}} \big( Q_t^{\text{veh}}, T_t \big).
\end{equation}
Due to heterogeneous capacities and training objectives, the infrastructure and vehicle models produce misaligned latent representations, with discrepancies in feature statistics and semantic embeddings. Direct fusion under such misalignment can introduce noise and degrade cooperative reasoning. To mitigate this issue, we employ a lightweight MLP-based adapter to project the infrastructure latent guidance into the vehicle latent space before fusion:
\begin{equation}
    \bar{z}_t^{\text{inf}}=\text{MLP}(z_t^{\text{inf}}).
\end{equation}
After alignment, an attention-based fusion mechanism then enables knowledge-aware latent integration:
\begin{equation}
\tilde{z}_{t}^{\text{veh}} = \text{CrossAttn} \big( z_t^{\text{veh}}, \bar{z}_t^{\text{inf}} \big), \quad
z_t^{\text{veh}}=\text{MLP}(\tilde{Q}_t^{\text{veh}}),
\end{equation} 
where $z_t^{\text{veh}}$ denotes the ego latent representation derived from vehicle semantic tokens. Through cross-attention, the ego vehicle selectively incorporates complementary global information, producing an enhanced cooperative semantic state $\tilde{z}_{t}^{\text{veh}}$ while remaining grounded in ego perception.

To further integrate the cooperative information into the reasoning process, we perform a recursive refinement via a second forward pass of the vehicle VLM:
\begin{equation}
\hat{z}_{t}^{\text{veh}} = \mathcal{F}_{\text{VLM}}^{\text{veh}} \big( \tilde{z}_{t}^{\text{veh}}, \mathcal{C}_t \big),
\end{equation}
where $\mathcal{C}_t=\{Q_t^{\text{veh}}, T_t\}$. 
This refinement step embeds the fused latent into the ego model’s internal reasoning dynamics, producing a cooperative semantic representation that reflects both global infrastructure guidance and local contextual understanding. $\hat{z}_{t}^{\text{veh}}$ is then utilized for downstream planning and trajectory generation, enabling informed and self-contained decision-making.

\begin{figure}[tb]
    \centering
    \includegraphics[width=1.0\linewidth]{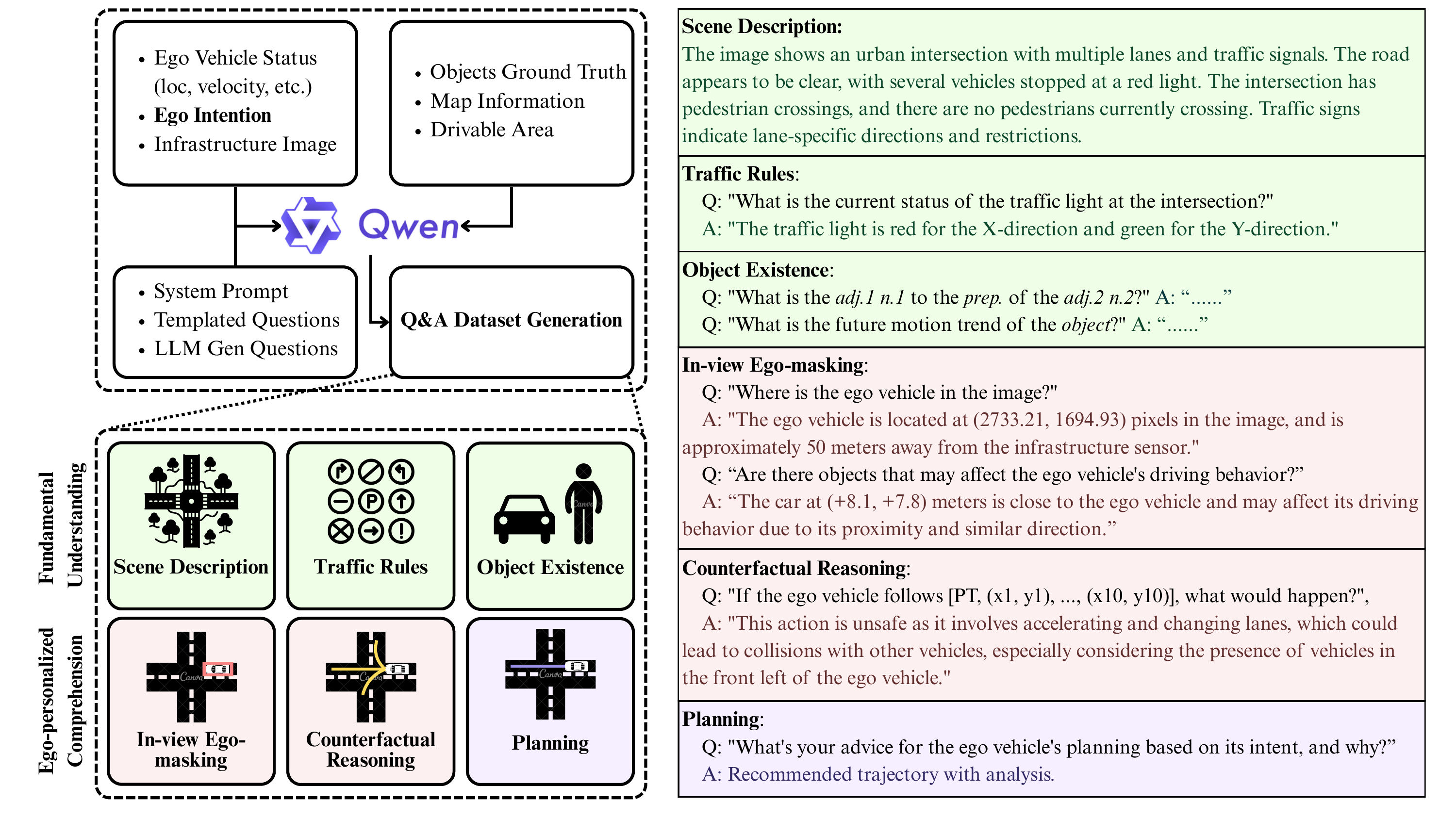}
    \caption{Construction and contents of the cooperation-oriented dataset.}
    \label{fig:dataset}
\end{figure}

\subsection{Cooperation-Oriented QA Dataset}
\label{sec:data}

To support and validate our dual-horizon cooperative framework, we introduce a cooperation-oriented QA benchmark that bridges the gap between sensor annotations and cooperative semantic reasoning. Specifically, this benchmark enhances the infrastructure’s semantic understanding capability, facilitating safer and more robust latent guidance for the ego vehicle. As illustrated in \cref{fig:dataset}, the dataset is generated through a two-stage pipeline with human-in-the-loop verification to ensure annotation quality and reasoning consistency. 

\textbf{Fundamental Scene Understanding.}
At this stage, we generate objective global scene descriptions that characterize traffic topology, lane constraints, and the spatial distribution and motion trends of surrounding agents. 
These annotations provide a consistent global context that facilitates downstream cooperative perception and decision-making.

\textbf{Ego-Personalized Comprehension.}
Beyond conventional QA datasets for single-intelligence autonomous driving, we explicitly model the auxiliary and strategic role of infrastructure intelligence in cooperative scenarios. Based on the fundamental stage, we align infrastructure reasoning with the ego vehicle by projecting the ego’s global pose into the infrastructure coordinate system, enabling explicit ego grounding in the global scene. Conditioned on this grounding, the infrastructure performs counterfactual reasoning over multiple what-if future trajectories generated under lane constraints and traffic rules to assess driving safety, interaction risks, and global traffic dynamics beyond the ego perception and prediction. Based on this information, the infrastructure outputs future-oriented safety assessments and planning-level strategic suggestions, such as potential collision risks, interaction conflicts, and recommended maneuver intentions. This ego-personalized reasoning further strengthens latent semantic guidance to support vehicle planning.

\section{Experiments}
\subsection{Experimental Setup}

The infrastructure system employs Qwen2.5-7B~\cite{qwen} to perform global and long-horizon semantic reasoning, while the vehicle system adopts Qwen2.5-0.5B~\cite{qwen} to meet on-board deployment constraints. The communication ratio is set to $5{:}1$, determined empirically based on computational time. Inference is conducted on NVIDIA GeForce RTX 3090 GPUs, and training is performed in multiple stages on NVIDIA A800 GPUs. Training consists of three stages. First, the infrastructure model is trained for scene understanding and ego-centric comprehension using visual data and our QA dataset with a learning rate of $\num{1e-5}$. Second, the ego model is trained using ego visual data with a learning rate of $\num{1e-4}$. Finally, the two systems are jointly optimized by training the HLA and IDLE for latent-level cooperation. All models are optimized using AdamW with a weight decay of 0.03. We select $\left| \mathcal{S} \right|=8$ intermediate layers for latent aggregation and set the temporal window size $k=5$.

The cooperation-oriented QA dataset is constructed on DAIR-V2X~\cite{dair}, a large-scale multi-modal benchmark for vehicle–infrastructure cooperative driving, containing approximately 100 scenes across 28 complex urban intersections. The question–answer pairs are generated using Qwen2.5-VL-32B~\cite{qwen-vl}, enabling semantically grounded supervision tailored to cooperative reasoning scenarios. The final dataset consists of over 80k QA pairs covering diverse cooperative driving scenarios. We further validate our framework on V2X-Sim~\cite{v2x-sim}, a simulated benchmark for cooperative autonomous driving.

\begin{table}[tb]
  \caption{Planning performance and safety evaluation.}
  \label{tab:planning}
  \centering
  \scriptsize
  \setlength{\tabcolsep}{3pt}
  \begin{tabular}{l|ccccc>{\columncolor{gray!20}}c|ccccc>{\columncolor{gray!20}}c}
    \toprule
    \multirow{2}{*}{Method} & \multicolumn{6}{c|}{L2 Error (\textit{m})$\downarrow$} & \multicolumn{6}{c|}{Collision Rate (\%)$\downarrow$} \\
     & 1\textit{s} & 2\textit{s} & 3\textit{s} & 4\textit{s} & 5\textit{s} & Avg. & 1\textit{s} & 2\textit{s} & 3\textit{s} & 4\textit{s} & 5\textit{s} & Avg. \\
    \midrule
    \multicolumn{13}{c}{Single-Intelligence Methods} \\
    \midrule
    VAD~\cite{vad} & 1.65 & 2.72 & 3.80 & 4.65 & 5.78 & 3.72 & 0.86 & 1.21 & 1.28 & 1.46 & 1.95 & 1.35 \\
    UniAD~\cite{uniad} & 1.26 & 2.22 & 3.06 & 3.97 & 5.22 & 3.15 & 0.88 & 1.18 & 1.32 & 1.04 & 1.45 & 1.17 \\
    SparseDrive~\cite{sparsedrive} & 1.02 & 1.69 & 2.37 & 2.99 & 3.87 & 2.39 & 0.46 & 1.23 & 1.28 & 1.34 & 1.58 & 1.18 \\
    OpenDriveVLA~\cite{opendrivevla} & 0.64 & 1.51 & 2.69 & 4.05 & 5.57 & 2.89 & 0.15 & 0.59 & 0.59 & 0.89 & \textbf{0.44} & 0.53 \\
    \midrule
    \multicolumn{13}{c}{Multi-Agent Cooperative Methods} \\
    \midrule
    V2VNet~\cite{v2vnet} & 1.96 & 2.37 & 3.41 & 3.91 & 4.73 & 3.28 & 0.74 & 0.88 & 1.03 & 1.24 & 1.58 & 1.09 \\
    CooperNaut~\cite{coopernaut} & 2.69 & 4.07 & 5.50 & 6.02 & 7.94 & 5.24 & 1.18 & 1.32 & 1.76 & 1.97 & 1.69 & 1.58 \\
    UniV2X~\cite{univ2x} & 1.45 & 2.19 & 3.04 & 3.96 & 5.24 & 3.18 & 0.15 & 0.15 & 0.44 & 0.59 & 0.74 & 0.41 \\
    UniMM-V2X~\cite{unimm-v2x} & 0.78 & 1.63 & 2.05 & 3.54 & 5.00 & 2.60 & \textbf{0.05} & 0.15 & 0.15 & 0.59 & 1.18 & 0.42 \\
    V2X-VLM~\cite{v2x-vlm} & 0.49 & 1.57 & 2.89 & 4.08 & 5.34 & 2.87 & \textbf{0.05} & \textbf{0.05} & 0.15 & 0.35 & 0.68 & 0.26 \\
    LangCoop~\cite{langcoop} & 0.30 & 0.93 & 1.96 & 3.17 & 4.58 & 2.19 & \textbf{0.05} & 0.15 & 0.44 & 0.87 & 1.02 & 0.51 \\
    \midrule
    \textbf{DH-VLM (Ours)} & \textbf{0.26} & \textbf{0.85} & \textbf{1.76} & \textbf{2.74} & \textbf{3.72} & \textbf{1.87} & \textbf{0.05} & \textbf{0.05} & \textbf{0.11} & \textbf{0.28} & \textbf{0.44} & \textbf{0.19} \\
    \bottomrule
  \end{tabular}
\end{table}

\begin{table}[tb]
    \centering
    \scriptsize
    \setlength{\tabcolsep}{13pt}
    \caption{On-board deployment, inference, and communication cost analysis.}
    \begin{tabular}{l|c|c|c}
    \toprule
    Method & \#Params (M)$\downarrow$ & Memory (GB)$\downarrow$ & Trans. Cost (BPS)$\downarrow$ \\
    \midrule
    V2VNet~\cite{v2vnet} & \textbf{262.1} & 6.09 & 8.19$\times10^7$ \\
    CooperNaut~\cite{coopernaut} & \textbf{262.1} & 6.09 & 8.19$\times10^7$ \\
    UniV2X~\cite{univ2x}  & 264.2 & 6.11 & 8.09$\times10^5$ \\
    UniMM-V2X~\cite{unimm-v2x} & 267.8 & 6.14 & 9.32$\times10^5$ \\
    \midrule
    V2X-VLM~\cite{v2x-vlm} & 776.5 & 10.92 & 1.24$\times10^7$ \\
    LangCoop~\cite{langcoop} & 7384.5 & 17.74 & \textbf{$\sim$ 4}{\boldmath$\times10^3$} \\
    \midrule
    DH-VLM (Ours) & 640.6 & \textbf{4.54} & 3.45$\times10^5$ \\
    \bottomrule
    \end{tabular}
    \label{tab:my_label}
\end{table}

\subsection{Main Results}

The planning results are reported in \cref{tab:planning}. We compare DH-VLM with single-agent baselines~\cite{vad, uniad, sparsedrive, opendrivevla} and cooperative methods~\cite{v2vnet, coopernaut, univ2x, unimm-v2x, v2x-vlm, langcoop}. DH-VLM achieves the best performance in both L2 error and collision rate. It reduces L2 error by \textbf{14.6\%} over \cite{langcoop} and lowers collision rate by \textbf{26.9\%} compared with \cite{v2x-vlm}, highlighting the benefit of latent-level cooperation. The improvement is more significant in long-horizon planning: relative to \cite{univ2x}, DH-VLM reduces L2 error by \textbf{29.0\%} and collision rate by \textbf{49.5\%}, highlighting the critical role of infrastructure-guided global reasoning in long-term decision making.

To assess deployment cost and communication efficiency, we compare model size, inference cost, and transmission cost across cooperative methods. Although DH-VLM does not reduce total parameter count compared with modular end-to-end frameworks~\cite{v2vnet, coopernaut, univ2x}, it lowers GPU memory usage by \textbf{25.5\%}, as runtime memory is dominated by intermediate activations rather than parameters. By replacing dense feature interaction with compact latent guidance, DH-VLM reduces activation and attention overhead, resulting in a reduced on-board inference footprint that better aligns with efficiency-constrained vehicles. Communication is required only at cooperation frames, where compact latent guidance is transmitted, leading to a \textbf{57.3\%} reduction compared with the query-based method~\cite{univ2x}. While \cite{langcoop} achieves lower bandwidth by sending natural language summaries, it is more sensitive to inaccurate cross-agent information. In contrast, latent-level communication preserves structured semantic guidance with improved robustness, as validated in \cref{sec:ablation}.

\begin{table}[tb]
    \centering
    \caption{Ablation study on the infrastructure system for ego-centric reasoning.}
    \scriptsize
    \setlength{\tabcolsep}{6.5pt}
    \begin{tabular}{c|cc|ccc>{\columncolor{gray!20}}c|ccc>{\columncolor{gray!20}}c}
    \toprule
    Fusion in  &  \multicolumn{2}{c|}{QA Level} & \multicolumn{4}{c|}{L2 Error ($m$)$\downarrow$} & \multicolumn{4}{c}{Collision Rate ($\%$)$\downarrow$} \\
    Vision Encoder & F & E & 3$s$ & 4$s$ & 5$s$ & Avg. & 3$s$ & 4$s$ & 5$s$ & Avg. \\
    \midrule
    - & - & - & 3.84 & 4.97 & 6.77 & 5.19 & 2.07 & 2.07 & 1.04 & 1.73 \\
    \checkmark & - & - & 2.37 & 3.86 & 5.52 & 3.92 & 0.59 & 1.19 & 1.30 & 1.03 \\
    \checkmark & \checkmark & - & 1.58 & 2.46 & 3.53 & 2.52 & \textbf{0.30} & 0.74 & 0.89 & 0.64 \\
    \checkmark & \checkmark & \checkmark & \textbf{0.99} & \textbf{1.67} & \textbf{2.51} & \textbf{1.72} & 0.44 & \textbf{0.44} & \textbf{0.44} & \textbf{0.44} \\
    \bottomrule
    \end{tabular}
    \begin{tablenotes}
    \scriptsize
    \item[] \textit{Note:} “F” and “E” denote fundamental scene understanding and ego-personalized comprehension component of our cooperation-oriented QA dataset, respectively. Results are directly evaluated from infrastructure-generated text outputs.
    \end{tablenotes}
    
    \label{tab:ablation-slow}
\end{table}

\begin{table}[tb]
    \centering
    \caption{Robustness of ego-vehicle planning under noisy infrastructure guidance.}
    \scriptsize
    \setlength{\tabcolsep}{3.3pt}
    \begin{tabular}{c|cc|c|ccccc>{\columncolor{gray!20}}c|ccccc>{\columncolor{gray!20}}c}
    \toprule
    \multirow{2}{*}{w./o.} & \multicolumn{2}{c|}{Level} & Inf & \multicolumn{6}{c|}{L2 Error ($m$)$\downarrow$} & \multicolumn{6}{c}{Collision Rate ($\%$)$\downarrow$} \\
     & T & L & Per & 1$s$ & 2$s$ & 3$s$ & 4$s$ & 5$s$ & Avg. & 1$s$ & 2$s$ & 3$s$ & 4$s$ & 5$s$ & Avg.\\
    \midrule
    - & - & - & - & 0.64 & 1.51 & 2.69 & 4.05 & 5.57 & 2.89 & 0.15 & 0.59 & 0.59 & 0.89 & \textbf{0.44} & 0.53 \\
    \midrule
    \checkmark & \checkmark & - & \scalebox{1.2}{$\bullet$} & \textbf{0.19} & \textbf{0.73} & \textbf{1.59} & \textbf{2.66} & \underline{3.98} & \textbf{1.83} & \textbf{0.05} & \textbf{0.05} & \textbf{0.05} & \textbf{0.25} & \underline{0.45} & \textbf{0.17} \\
    \checkmark & \checkmark & - & \scalebox{1.2}{$\circ$} & 0.44 & 1.26 & 2.31 & 3.42 & 4.98 & 2.48 & 0.15 & 0.25 & 0.34 & 0.69 & 1.27 & 0.54 \\
    \checkmark & \checkmark & - & \scalebox{0.75}{$\triangle$} & 2.26 & 3.84 & 5.63 & 7.40 & 9.15 & 5.66 & 0.43 & 0.75 & 1.84 & 3.87 & 4.99 & 2.38 \\
    \midrule
    \checkmark & - & \checkmark & \scalebox{1.2}{$\bullet$} & \underline{0.26} & \underline{0.85} & \underline{1.76} & \underline{2.74} & \textbf{3.72} & \underline{1.87} & \textbf{0.05} & \textbf{0.05} & \underline{0.11} & \underline{0.28} & \textbf{0.44} & \underline{0.19} \\
    \checkmark & - & \checkmark & \scalebox{1.2}{$\circ$} & 0.24 & 0.94 & 1.82 & 2.91 & 3.99 & 1.98 & 0.05 & 0.15 & 0.24 & 0.36 & 0.58 & 0.28 \\
    \checkmark & - & \checkmark & \scalebox{0.75}{$\triangle$} & 0.26 & 0.95 & 2.07 & 3.17 & 4.50 & 2.19 & 0.15 & 0.25 & 0.57 & 0.79 & 1.02 & 0.56 \\
    \bottomrule
    \end{tabular}
    \begin{tablenotes}
    \scriptsize
    \item[] \textit{Note:}  “T” and “L” denote text-level and latent-level guidance, respectively. 
    \item[] \scalebox{1.2}{$\bullet$}, \scalebox{1.2}{$\circ$}, and \scalebox{0.75}{$\triangle$} indicate average 5s L2 error of infrastructure-generated trajectories ($<2$m, $2\sim4$m, $>4$m), indicating the infrastructure’s latent reasoning capability for the ego vehicle.
    \end{tablenotes}
    \label{tab:ablation-fast}
\end{table}

\subsection{Ablation Study}
\label{sec:ablation}

\textbf{Performance of the Infrastructure in Assisting the Vehicle.} As shown in \cref{tab:ablation-slow}, the Fusion Vision Encoder is essential for holistic scene understanding at the infrastructure side. Since the ego vehicle may lie partially or entirely outside the infrastructure’s field of view, ego-centric reasoning is infeasible without multi-agent feature alignment and fusion. Without this fusion, the infrastructure often produces incomplete or inaccurate guidance, which can mislead ego planning. We further examine the effect of different components in the cooperation-oriented QA dataset. The Fundamental Scene Understanding data provides global traffic context and spatial structure, while the Ego-Personalized Comprehension data explicitly connects global perception to ego-centric decision-making and long-horizon safety reasoning. These findings indicate that both multi-agent visual fusion and structured cooperative QA supervision are critical for reliable global reasoning and high-quality infrastructure guidance.

\begin{table}[tb]
    \centering
    \caption{Impact of different communication ratios and the number of selected intermediate layers in HLA for latent aggregation on driving performance.}
    \scriptsize
    \setlength{\tabcolsep}{3.5pt}
    \begin{tabular}{c|c|ccccc>{\columncolor{gray!20}}c|ccccc>{\columncolor{gray!20}}c}
    \toprule
    Comm. & \# Fused & \multicolumn{6}{c|}{L2 Error ($m$)$\downarrow$} & \multicolumn{6}{c}{Collision Rate ($\%$)$\downarrow$} \\
    Ratio & Layers$^\dagger$ & 1$s$ & 2$s$ & 3$s$ & 4$s$ & 5$s$ & Avg. & 1$s$ & 2$s$ & 3$s$ & 4$s$ & 5$s$ & Avg.\\
    \midrule
    \rowcolor{gray!20}$5{:}1$ & 8 & \underline{0.26} & 0.85 & \underline{1.76} & \underline{2.74} & \textbf{3.72} & \underline{1.87} & \underline{0.05} & \textbf{0.05} & \underline{0.11} & \underline{0.28} & \underline{0.44} & \underline{0.19} \\
    $5{:}1$ & 16 & 0.28 & \underline{0.84} & 1.79 & 2.75 & 3.74 & 1.88 & \underline{0.05} & \textbf{0.05} & 0.15 & 0.35 & 0.55 & 0.23 \\
    $5{:}1$ & 4 & 0.33 & 0.91 & 1.86 & 2.83 & \underline{3.73} & 1.93 & \underline{0.05} & 0.15 & 0.15 & 0.35 & 0.60 & 0.26 \\
    $5{:}1$ & 1 & 0.45 & 0.97 & 2.01 & 2.97 & 3.91 & 2.06 & 0.15 & 0.15 & 0.25 & 0.45 & 0.58 & 0.32 \\
    \midrule
    $10{:}1$ & 8 & 0.36 & 1.02 & 2.18 & 2.95 & 4.56 & 2.21 & \underline{0.05} & 0.15 & 0.35 & 0.35 & 0.53 & 0.29 \\
    $10{:}1$ & 1 & 0.53 & 1.25 & 2.35 & 3.87 & 4.92 & 2.58 & 0.15 & 0.35 & 0.48 & 0.55 & 0.75 & 0.46 \\
    \midrule
    $1{:}1$* & 8 & \textbf{0.23} & \textbf{0.79} & \textbf{1.68} & \textbf{2.57} & \underline{3.73} & \textbf{1.80} & \textbf{0.03} & \textbf{0.05} & \textbf{0.05} & \textbf{0.19} & \textbf{0.33} & \textbf{0.13} \\
    $1{:}1$* & 1 & 0.35 & 0.98 & 1.97 & 2.95 & 3.87 & 2.02 & 0.10 & \underline{0.10} & 0.25 & \underline{0.28} & 0.67 & 0.28 \\
    \bottomrule
    \end{tabular}
    \begin{tablenotes}
    \scriptsize
    \item[] $\dagger$ Fused layer number = 1 denotes using the final hidden layer directly.
    \item[*] * Theoretical setting beyond practical deployment constraints.
    \end{tablenotes}
    \label{tab:ablation-freq-fused}
\end{table}
\begin{table}[h]
    \centering
    \scriptsize
    \setlength{\tabcolsep}{33.7pt}
    \caption{Ablation study on module effectiveness. 'w/o' denotes replacing the corresponding module with an MLP-based variant.}
    \begin{tabular}{l|c|c}
    \toprule
    Method & L2 Avg. (m)$\downarrow$ & CR Avg. (\%)$\downarrow$\\
    \midrule
    DH-VLM (Full) & \textbf{1.87} & \textbf{0.19} \\
    \midrule
    w/o HLA & 2.31 & 0.39 \\
    w/o IDLE & 2.48 & 0.34 \\
    w/o both & 2.59 & 0.47 \\
    \bottomrule
    \end{tabular}
    \label{tab:module}
\end{table}

\textbf{Robustness of Infrastructure Guidance.} As shown in \cref{tab:ablation-fast}, although the infrastructure possesses stronger global reasoning capability, its text-based guidance $T_t^{\text{inf}}$ inevitably contains noise due to long-horizon uncertainty and asynchronous sensing. Direct injection propagates these errors to the ego planning process, sometimes degrading performance beyond single-agent baselines (+95.8\% L2 error, +349.1\% collision rate), indicating that naive cooperation can be harmful. In contrast, our latent-level fusion treats infrastructure guidance as a soft prior, allowing the ego model to selectively absorb useful semantic cues while suppressing unreliable signals. Notably, it achieves performance comparable to the best text-level fusion, with improved robustness and safety.

\textbf{Communication Frequency and Layer Selection.}
As shown in \cref{tab:ablation-freq-fused}, a 1:1 communication ratio yields the best performance, but it is impractical for deployment. Aligning with the inference speeds of the systems, a 5:1 ratio achieves a better balance between communication cost and performance, while lower frequencies under-utilize infrastructure assistance. For latent aggregation, increasing the number of intermediate layers brings marginal gains due to redundancy, whereas too few layers provide insufficient semantic richness. Notably, transmitting only the final layer causes significant semantic degradation, as it is already close to text decoding and diverges from the visual semantic space, resulting in substantial information loss.

\textbf{Component Analysis.}
We analyze HLA and IDLE by replacing each with a simple MLP while keeping the remaining architecture unchanged. As shown in \cref{tab:module}, replacing HLA increases the L2 error by 23.5\% and doubles the collision rate, highlighting the role of historical latent aggregation in robust cooperative planning. Replacing IDLE increases the L2 error by 32.6\% and the collision rate by 78.9\%, demonstrating the importance of infrastructure-guided latent refinement for accurate trajectory generation. These observations highlight the complementary roles of HLA and IDLE and attribute the performance gains to the proposed latent guidance instead of increased model complexity.

\begin{figure}[tb]
    \centering
    \begin{subfigure}[b]{0.3\textwidth}
        \centering
        \includegraphics[width=\textwidth]{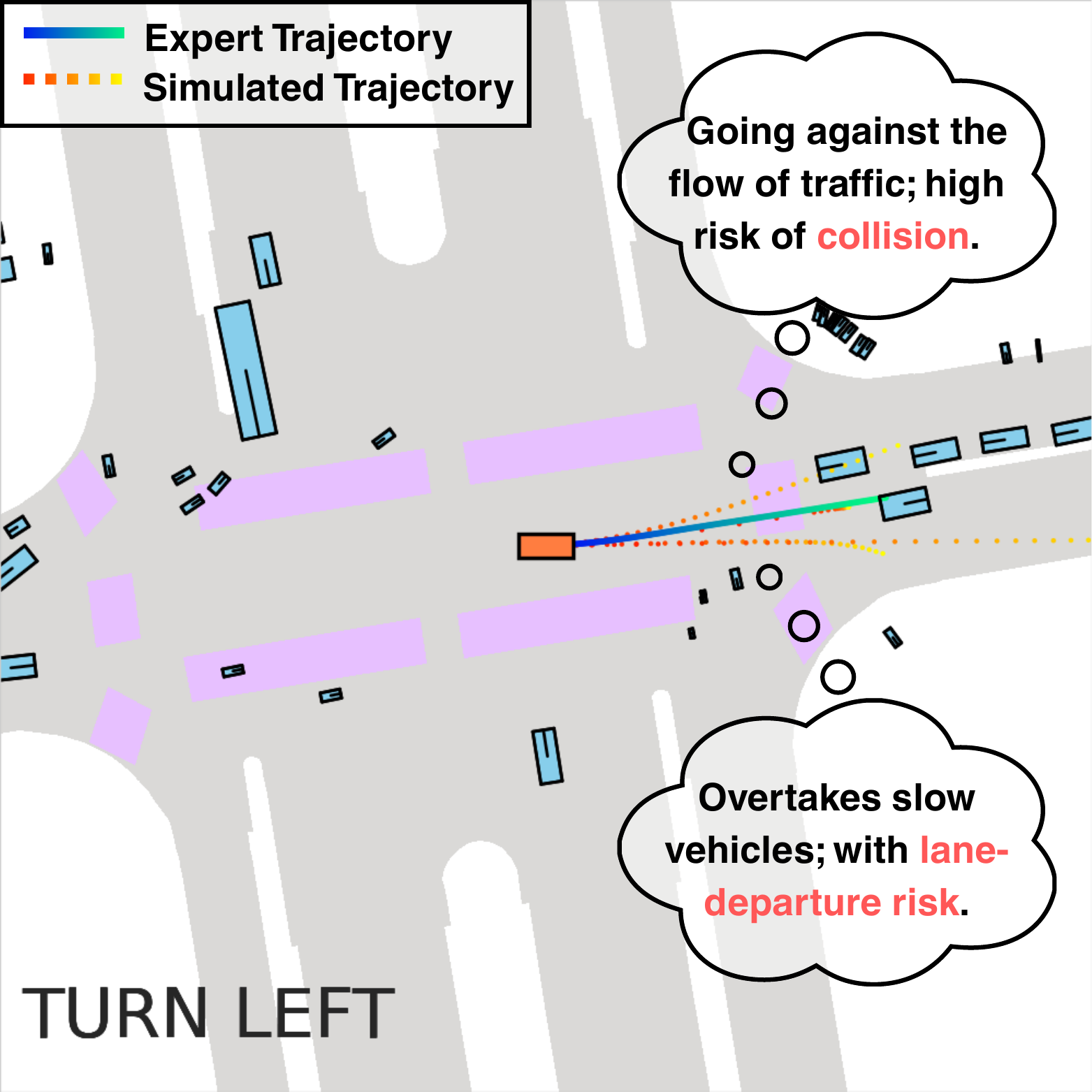}
        \caption{}
        \label{fig:counterfactual-left}
    \end{subfigure}
    \hspace{1em}
    \begin{subfigure}[b]{0.3\textwidth}
        \centering
        \includegraphics[width=\textwidth]{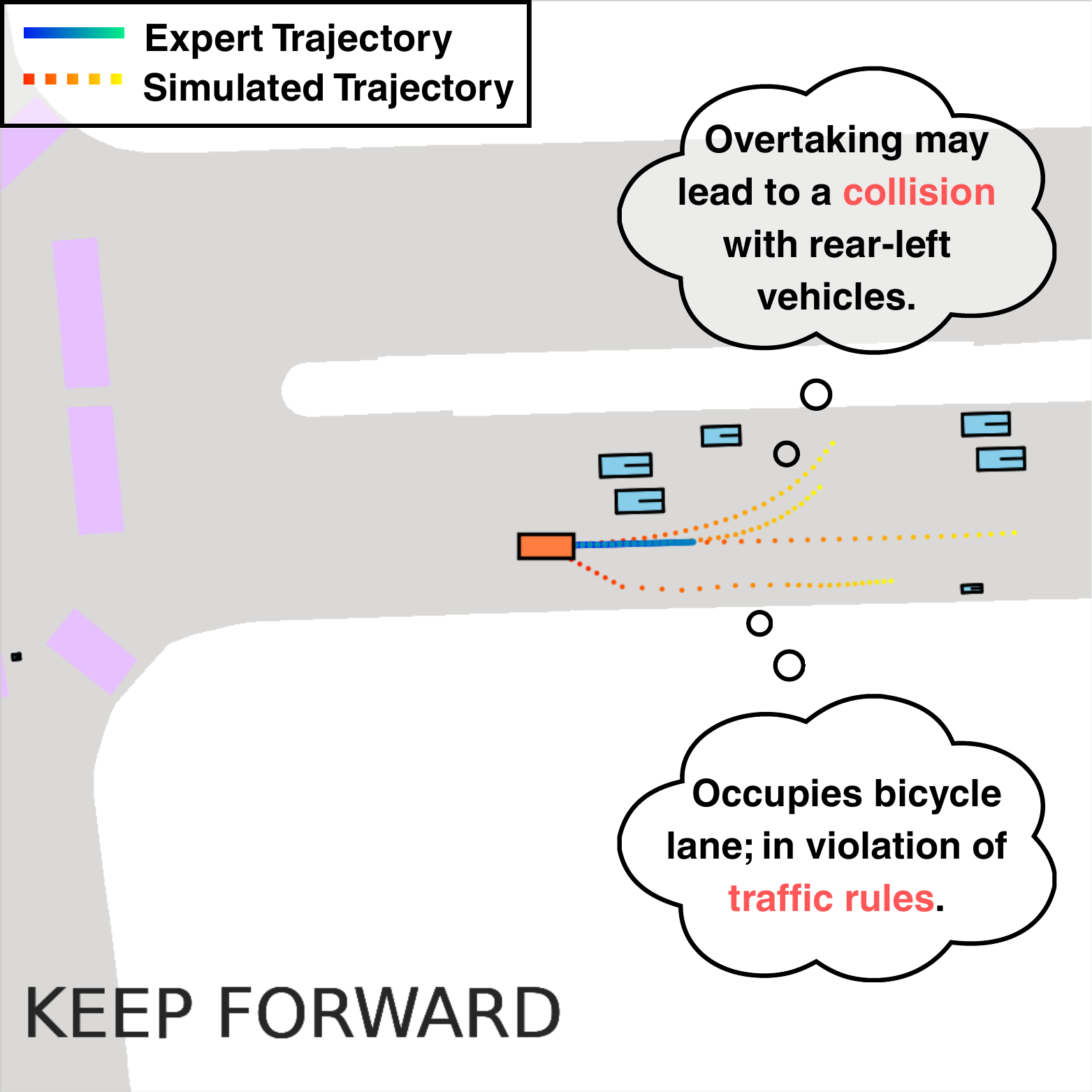}
        \caption{}
        \label{fig:counterfactual-forward}
    \end{subfigure}
    \hspace{1em}
    \begin{subfigure}[b]{0.3\textwidth}
        \centering
        \includegraphics[width=\textwidth]{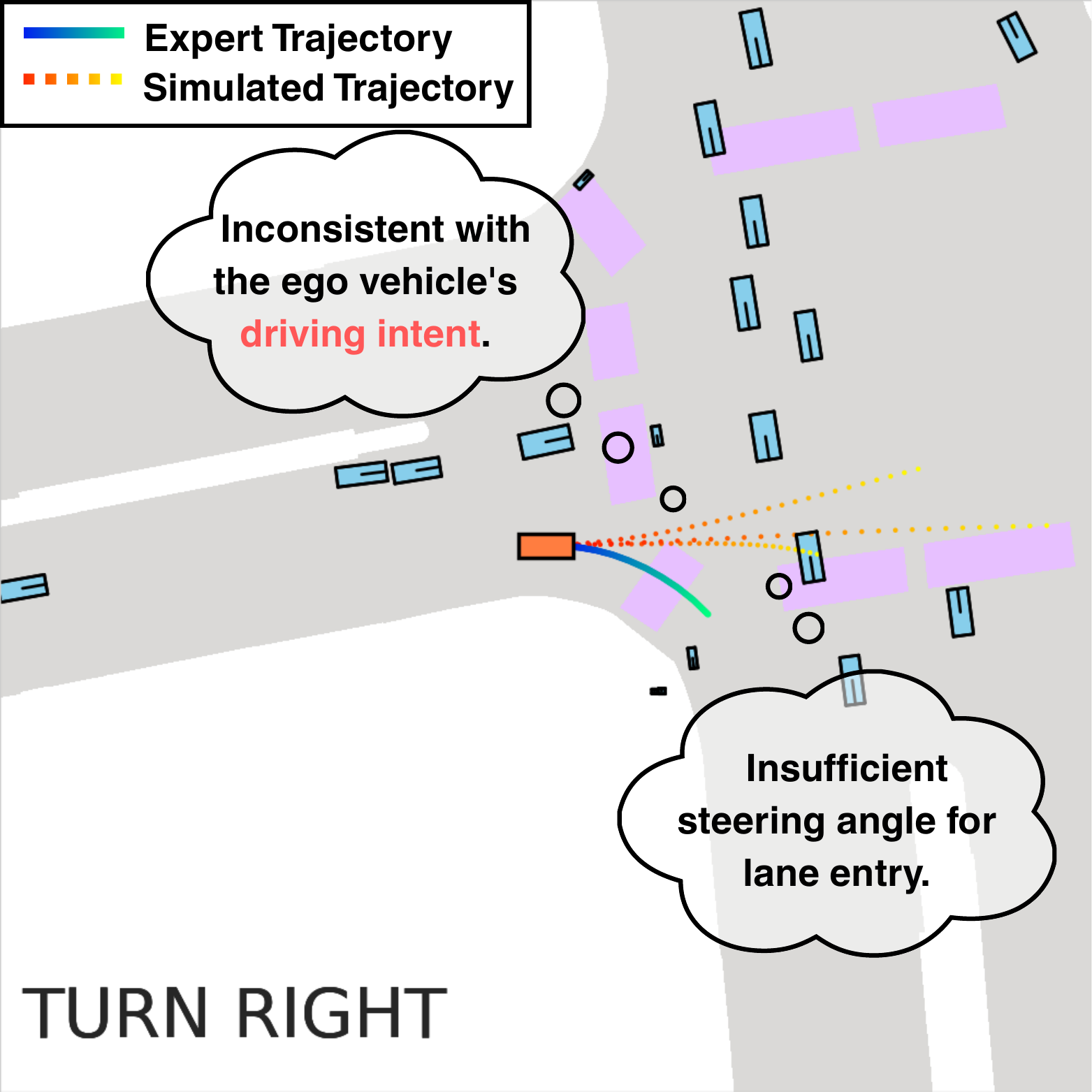}
        \caption{}
        \label{fig:counterfactual-right}
    \end{subfigure}
    \caption{Counterfactual reasoning conducted by the infrastructure.}
    \label{fig:counterfactual}
\end{figure}

\begin{figure}[tb]
    \centering
    \begin{subfigure}[b]{0.3\textwidth}
        \centering
        \includegraphics[width=\textwidth]{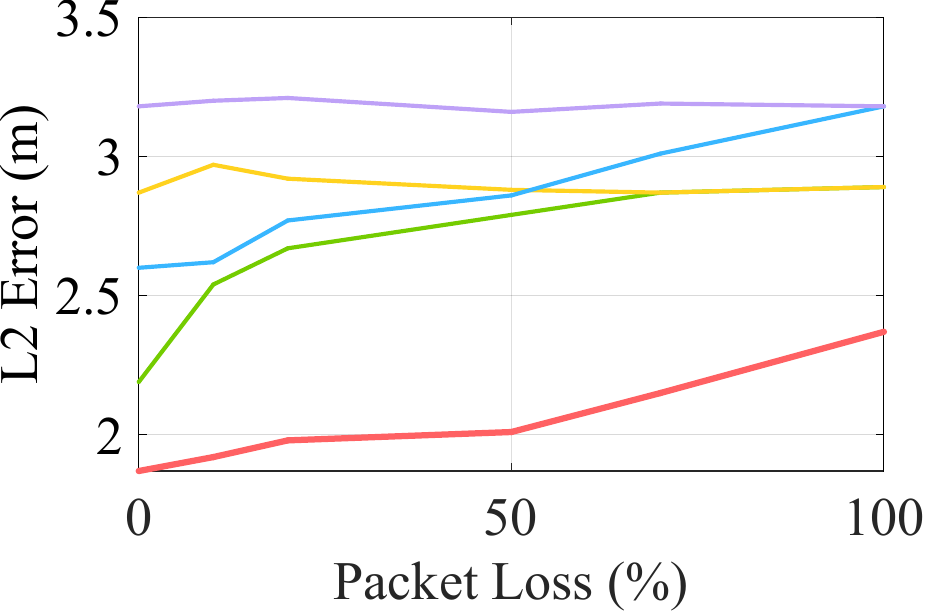}
        \label{fig:package_loss}
    \end{subfigure}
    \hspace{3em}
    \begin{subfigure}[b]{0.3\textwidth}
        \centering
        \includegraphics[width=\textwidth]{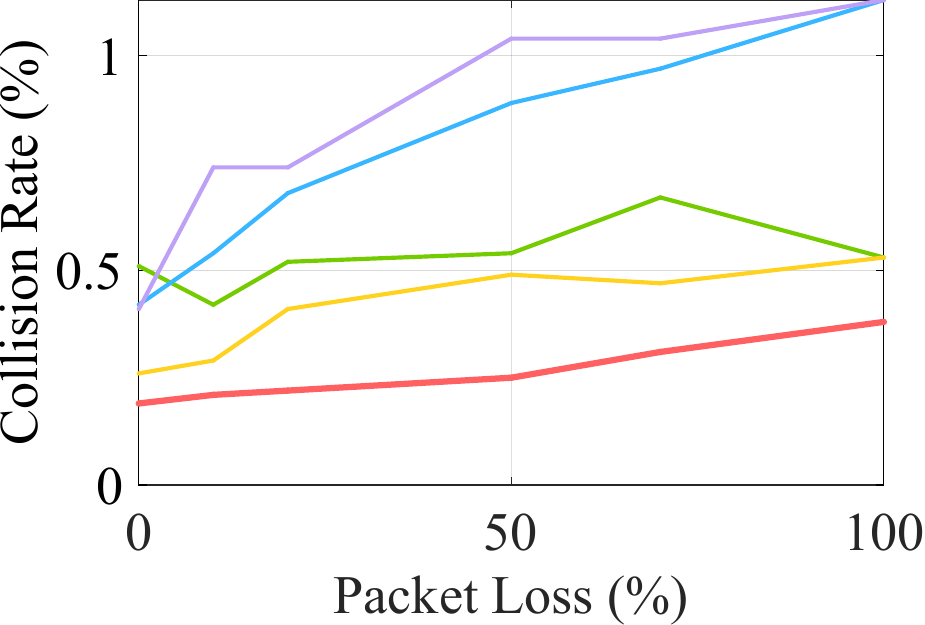}
        \label{fig:package_loss_colli}
    \end{subfigure}
    \hspace{3em}
    \begin{subfigure}[b]{0.3\textwidth}
        \centering
        \includegraphics[width=\textwidth]{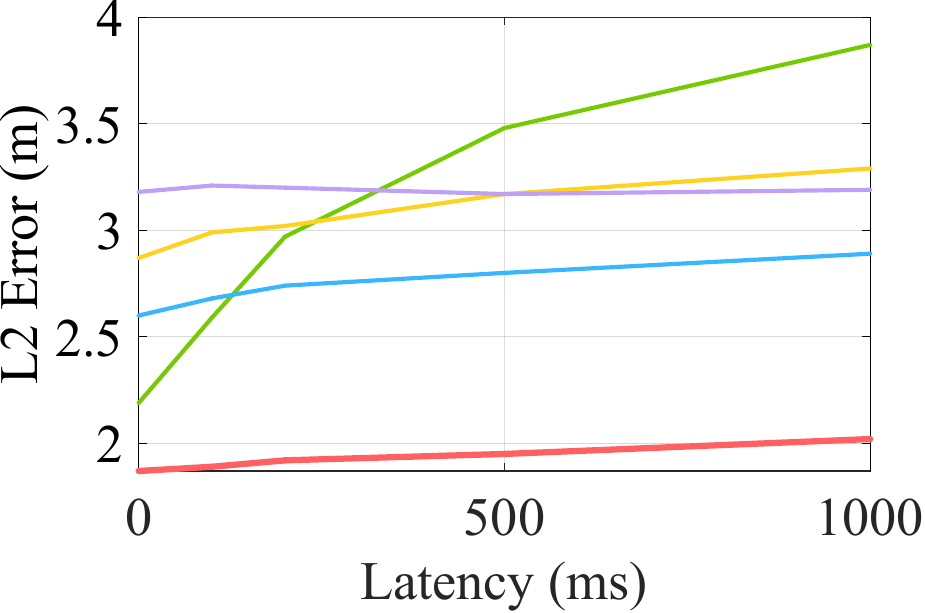}
        \label{fig:latency}
    \end{subfigure}
    \hspace{3em}
    \begin{subfigure}[b]{0.3\textwidth}
        \centering
        \includegraphics[width=\textwidth]{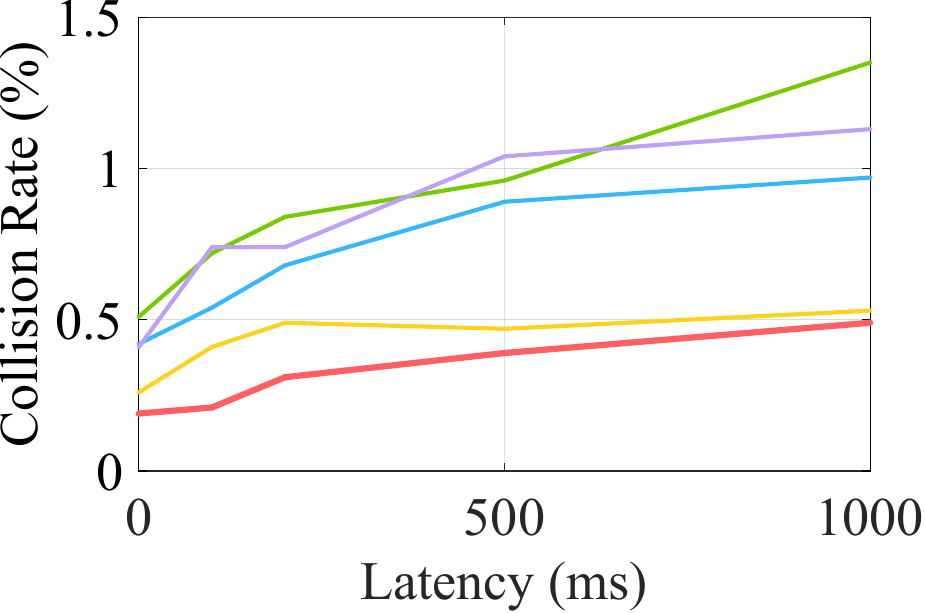}
        \label{fig:latency_colli}
    \end{subfigure}
    \caption{Communication robustness experiments. Methods: \textcolor{RoseRed}{DH-VLM (ours)}, \textcolor{GrassGreen}{LangCoop}, \textcolor{EggYellow}{V2X-VLM}, \textcolor{SkyBlue}{UniMM-V2X}, \textcolor{FlowerPurple}{UniV2X}.}
    \label{fig:communication_stability}
\end{figure}

\begin{figure}[tb]
    \centering
    \includegraphics[width=\linewidth]{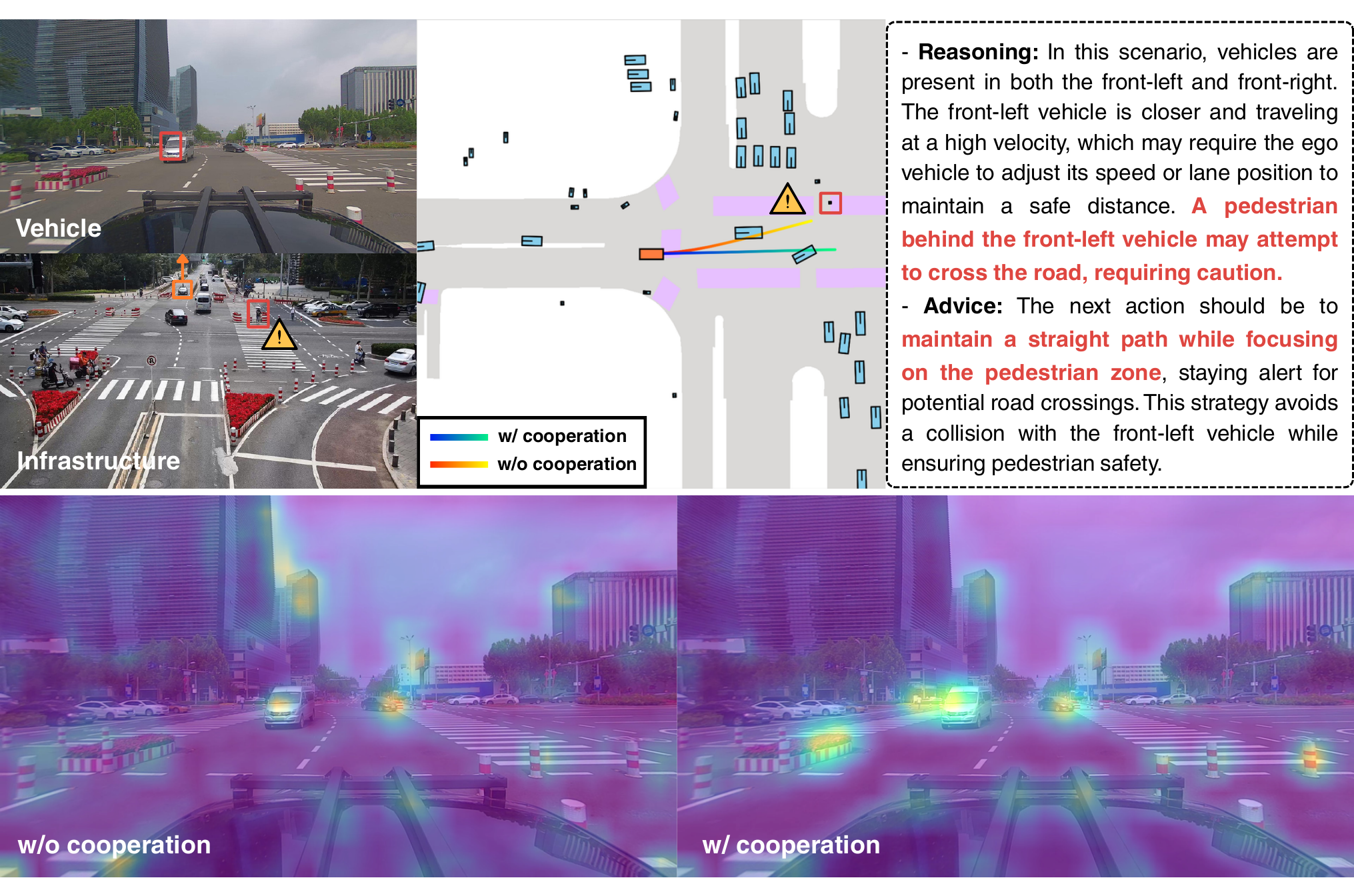}
    \caption{Cooperative driving results and latent-derived spatial activations of DH-VLM, with infrastructure textual reasoning shown for interpretability.}
    \label{fig:planning_result}
\end{figure}

\subsection{System Stability and Degraded Communication Analysis}

To evaluate robustness under realistic communication conditions, we analyze DH-VLM under varying packet loss rates and communication latencies. As shown in \cref{fig:communication_stability}, DH-VLM maintains stable planning performance across a wide range of communication degradations. This robustness is enabled by asynchronous communication and temporal latent compensation. Specifically, DH-VLM adopts a sparse 5:1 communication ratio, reducing bandwidth requirements and relaxing synchronization constraints. Instead of transmitting raw sensor data~\cite{v2x-vlm}, intermediate features~\cite{univ2x,unimm-v2x}, or text-based messages~\cite{langcoop}, latent-level communication exchanges compact semantic representations, improving resilience to communication impairments while mitigating performance degradation and safety risks, further validating the infrastructure guidance robustness analysis in \cref{sec:ablation}. When infrastructure guidance is delayed or unavailable, e.g., under 100\% packet loss or latencies exceeding 500ms, the ego vehicle operates independently using its onboard model, enabling stable asynchronous cooperation.

\subsection{Qualitative Visualization}

\textbf{Ego-Personalized Comprehension.} To qualitatively evaluate the proposed infrastructure-assisted reasoning mechanism, we visualize the ego-centric counterfactual reasoning process in \cref{fig:counterfactual}. In left-turn, forward-driving, and right-turn scenarios, the infrastructure generates multiple candidate trajectories and selects the safest one conditioned on the intended maneuver of the ego vehicle. The chosen trajectories consistently follow traffic rules and avoid potential collision regions. These observations suggest that the counterfactual reasoning module enables intention-aware safety assessment beyond reactive collision avoidance. By reasoning over alternative futures conditioned on ego intent, the infrastructure gains a clearer understanding of complex traffic interactions and potential risks, thereby supporting more reliable cooperative planning decisions.

\textbf{Latent Evolution and Planning under Cooperation.} 
\cref{fig:planning_result} presents qualitative planning results of DH-VLM, along with infrastructure-generated textual outputs for interpretability. 
To analyze how infrastructure guidance affects ego latent evolution, we also visualize spatial activation maps by projecting ego latent representations through a lightweight perception head. In heavily occluded intersections, the ego-only model may fail to detect pedestrians hidden behind foreground vehicles. With infrastructure support, the latent representations encode a more complete and globally consistent scene understanding, enabling the ego vehicle to anticipate pedestrian motion and adjust its trajectory. The spatial activation maps show more concentrated and semantically structured responses around critical regions, particularly near occluded pedestrians and interaction areas, suggesting that the latent space captures safety-critical contextual cues that are difficult to infer from ego-only perception.

\section{Conclusion}

This paper presents DH-VLM, a dual-horizon cooperative driving architecture that assigns global-reasoning horizon understanding to the infrastructure while preserving local-planning horizon at the vehicle. To bridge heterogeneous computational capacities, DH-VLM enables latent-level cooperation, allowing the infrastructure to provide semantic guidance with high communication efficiency and robustness while minimizing the impact of communication noise and imperfect infrastructure guidance on ego vehicle's driving decisions. In addition, we construct a cooperation-oriented QA dataset, comprising fundamental understanding and ego-personalized comprehension tasks to explicitly enhance the system’s ability to reason for the ego vehicle. Extensive experiments demonstrate that DH-VLM achieves state-of-the-art performance. With a 57.3\% reduction in communication bandwidth, our method reduces the 5s L2 planning error by 14.6\% and the collision rate by 26.9\%. Notably, the ego vehicle only requires a lightweight onboard model, significantly improving the practical deployability of cooperative autonomous driving systems. However, due to the lack of standardized closed-loop evaluation methods for cooperative driving, our current experiments are limited to open-loop evaluation. Moreover, the present framework remains within the VLM paradigm. Future work will extend the approach toward a VLA-based cooperative architecture.

\clearpage

%
%
\bibliographystyle{splncs04}
\bibliography{main}
\end{document}